\pdfoutput=1

\documentclass[11pt]{article}

\usepackage[]{emnlp2021}

\usepackage{times}
\usepackage{latexsym}

\usepackage{microtype}
\usepackage{graphicx}
\usepackage{subfigure}
\usepackage{booktabs} 

\usepackage{verbatim}
\usepackage{comment}
\usepackage{caption}
\usepackage{wrapfig}
\usepackage{tablefootnote}
\usepackage{url}
\usepackage{array,multirow}
\usepackage{float}
\usepackage{textcomp}
\usepackage{nccmath}
\usepackage{relsize}
\usepackage{makecell}
\usepackage[ruled]{algorithm2e}
\usepackage{color}
\usepackage[export]{adjustbox}
\usepackage{kotex} 
\usepackage{amsmath,amssymb,amsfonts,amsthm,mathtools,bm}
\usepackage{arydshln}
\usepackage{enumitem}

\usepackage{hyperref} 
\hypersetup{colorlinks=true}

\usepackage[T1]{fontenc}

\usepackage[utf8]{inputenc}

\usepackage{microtype}

\definecolor{ouryellow}{rgb}{0.5, 0.85, 0.4}
\definecolor{ourred}{rgb}{1.0, 0.8, 0.8}
\definecolor{ourgreen}{rgb}{0.50, 0.78, 0.81}
\definecolor{ourorange}{rgb}{0.94, 0.61, 0.38}

\def\myeqref#1{Eq.~\eqref{#1}}

\title{Distilling Linguistic Context for Language Model Compression}

\author{Geondo Park$^{1}$ \qquad Gyeongman Kim$^{1}$ \qquad Eunho Yang$^{1,2}$ \\
  KAIST$^{1}$, Daejeon, South Korea \\
  AITRICS$^{2}$, Seoul, South Korea \\ 
  \texttt{\{geondopark, gmkim, eunhoy\}@kaist.ac.kr}\\}

\begin{document}
\maketitle
\begin{abstract}
A computationally expensive and memory intensive neural network lies behind the recent success of language representation learning. Knowledge distillation, a major technique for deploying such a vast language model in resource-scarce environments, transfers the knowledge on individual word representations learned without restrictions. In this paper, inspired by the recent observations that language representations are relatively positioned and have more semantic knowledge as a whole, we present a new knowledge distillation objective for language representation learning that transfers the \emph{contextual} knowledge via two types of relationships across representations: \emph{Word Relation} and \emph{Layer Transforming Relation}. Unlike other recent distillation techniques for the language models, our contextual distillation does not have any restrictions on architectural changes between teacher and student. We validate the effectiveness of our method on challenging benchmarks of language understanding tasks, not only in architectures of various sizes, but also in combination with DynaBERT, the recently proposed adaptive size pruning method. 

\end{abstract}

\section{Introduction}
Since the Transformer, a simple architecture based on attention mechanism, succeeded in machine translation tasks, Transformer-based models have become a new state of the arts that takes over more complex structures based on recurrent or convolution networks on various language tasks, e.g., language understanding and question answering, etc~\cite{bert, albert, roberta, raffel2019exploring, xlnet}. However, in exchange for high performance, these models suffer from a major drawback: tremendous computational and memory costs. In particular, it is not possible to deploy such large models on platforms with limited resources such as mobile and wearable devices, and it is an urgent research topic with impact to keep up with the performance of the latest models from a \emph{small-size} network. 

As the main method for this purpose, Knowledge Distillation~(KD) transfers knowledge from the large and well-performing network (teacher) to a smaller network (student). There have been some efforts that distill Transformer-based models into compact networks~\cite{distilbert, pd, pkd, mobilebert, tinybert, minilm}.

However, they all build on the idea that each word representation is independent, ignoring relationships between words that could be more informative than individual representations.

In this paper, we pay attention to the fact that word representations from language models are very \emph{structured} and capture certain types of semantic and syntactic relationships. - Word2Vec~\cite{wor2vec} and Glove~\cite{glove} demonstrated that trained embedding of words contains the linguistic patterns as linear relationships between word vectors. Recently, \citet{manning} found out that the distance between words contains the information of the dependency parse tree. Many other studies also suggested the evidence that contextual word representations~\cite{belinkov2017neural, tenney2019bert, tenney2019you} and attention matrices~\cite{vig2019visualizing, clark2019does} contain important relations between words. Moreover, \citet{identifiability} showed the vertical relations in word representations across the transformer layers through word identifiability. Intuitively, although each word representation has respective knowledge, the set of representations of words as a whole is more semantically meaningful, since  words in the embedding space are positioned relatively by learning.

Inspired by these observations, we propose a novel distillation objective, termed Contextual Knowledge Distillation~(CKD), for language tasks that utilizes the statistics of relationships between word representations. In this paper, we define two types of contextual knowledge: \emph{Word Relation~(WR)} and \emph{Layer Transforming Relation~(LTR)}. Specifically, WR is proposed to capture the knowledge of \emph{relationships between word representations} and LTR defines \emph{how each word representation changes} as it passes through the network layers. 

We validate our method on General Language Understanding Evaluation (GLUE) benchmark and the Stanford Question Answer Dataset~(SQuAD), and show the effectiveness of CKD against the current state-of-the-art distillation methods. To validate elaborately, we conduct experiments on task-agnostic and task-specific distillation settings. We also show that our CKD performs effectively on a variety of network architectures. Moreover, with the advantage that CKD has no restrictions on student's architecture, we show CKD further improves the performance of adaptive size pruning method \cite{dynabert} that involves the architectural changes during the training.

To summarize, our contribution is threefold:
\begin{itemize}[leftmargin=8mm]
\item (1) Inspired by the recent observations that word representations from neural networks are structured, we propose a novel knowledge distillation strategy, Contextual Knowledge Distillation (CKD), that transfers the relationships across word representations. 
\item (2) We present two types of complementary contextual knowledge: horizontal Word Relation across representations in a single layer and vertical Layer Transforming Relation across representations for a single word. 
\item (3) We validate CKD on the standard language understanding benchmark datasets and show that CKD not only outperforms the state-of-the-art distillation methods but boosts the performance of adaptive pruning method.
\end{itemize}

\section{Related Work}

\paragraph{Knowledge distillation} Since recently popular deep neural networks are computation- and memory-heavy by design, there has been a long line of research on transferring knowledge for the purpose of compression. \citet{hinton2015distilling} first proposed a teacher-student framework with an objective that minimizes the KL divergence between teacher and student class probabilities. In the field of natural language processing (NLP), knowledge distillation has been actively studied \cite{seq-kd, hu2018attention}. In particular, after the emergence of large language models based on pre-training such as BERT~\citep{bert, roberta, xlnet, raffel2019exploring}, many studies have recently emerged that attempt various knowledge distillation in the pre-training process and/or fine-tuning for downstream tasks in order to reduce the burden of handling large models. Specifically, \citet{tang2019distilling, chia2019transformer} proposed to distill the BERT to train the simple recurrent and convolution networks. \citet{distilbert, pd} proposed to use the teacher's predictive distribution to train the smaller BERT and \citet{pkd} proposed a method to transfer individual representation of words. In addition to matching the hidden state, \citet{tinybert, mobilebert, minilm} also utilized the attention matrices derived from the Transformer. Several works including~\citet{liu2020fastbert, dynabert} improved the performance of other compression methods by integrating with knowledge distillation objectives in the  training procedure. In particular, DynaBERT~\cite{dynabert} proposed the method to train the adaptive size BERT using the hidden state matching distillation. Different from previous knowledge distillation methods that transfer respective knowledge of word representations, we design the objective to distill the contextual knowledge contained among word representations.

\paragraph{Contextual knowledge of word representations} Understanding and utilizing the relationships across words is one of the key ingredients in language modeling. Word embedding \citep{wor2vec,glove} that captures the context of a word in a document, has been traditionally used. Unlike the traditional methods of giving fixed embedding for each word, the contextual embedding methods \citep{bert, ELMo} that assign different embeddings according to the context with surrounding words have become a new standard in recent years showing high performance. \citet{sentiment_classification} improved the performance of the sentiment classification task by using word relation, and \citet{probe_structural,manning} found that the distance between contextual representations contains syntactic information of sentences. Recently, \citet{identifiability} also experimentally showed that the contextual representations of each token change over the layers.
Our research focuses on knowledge distillation using context information between words and between layers, and to our best knowledge, we are the first to apply this context information to knowledge distillation.

\section{Setup and background}\label{sec:setup}

Most of the recent state-of-the-art language models are stacking Transformer layers which consist of repeated multi-head attentions and position-wise feed-forward networks.

\paragraph{Transformer based networks.}

Given an input sentence with $n$ tokens, $\bm{X}= [x_{1}, x_{2},\dots,x_{n}] \in \mathbb{R}^{d_i \times n}$, most networks~\citep{bert, albert, roberta} utilize the embedding layer to map an input sequence of symbol representations $\bm{X}$ to a sequence of continuous representations $\bm{E} = [e_{1}, \dots, e_{n}] \in \mathbb{R}^{d_e \times n}$. Then, each $l$-th Transformer layer of the identical structure takes the previous representations $\bm{R_{l-1}}$ and produces the updated representations $\bm{R_{l}}=[r_{l,1}, r_{l,2}, \dots, r_{l,n}] \in \mathbb{R}^{d_r \times n}$ through two sub-layers: Multi-head Attention (MHA) and position-wise Feed Forward Network~(FFN). The input at the first layer~($l=1$) is simply $\bm{E}$. In MHA operation where $h$ separate attention heads are operating independently, each input token $r_{l-1,i}$ for each head is projected into a query $q_i \in \mathbb{R}^{d_q}$, key $k_i \in \mathbb{R}^{d_q}$, and value $v_i \in \mathbb{R}^{d_v}$, typically $d_k = d_q = d_v=d_r/h$. Here, the key vectors and value vectors are packed into the matrix forms $\bm{K} =[k_1,\cdots,k_{n}]$ and $\bm{V} =[v_1,\cdots,v_{n}]$, respectively, and the attention value $a_i$ and output of each head $o_{h,i}$ are calculated as followed:
\begin{equation}
    a_i = \textrm{Softmax}\left(\frac{\bm{K}^T \cdot q_i}{\sqrt{d_q}}\right)  \quad \text{and } \quad o_{h,i} = \bm{V} \cdot a_i \nonumber
\end{equation}

The outputs of all heads are then concatenated and fed through the FFN, producing the single word representation $r_{l,i}$.
For clarity, we pack attention values of all words into a matrix form $\bm{A_{l,h}} = [a_1, a_2, .., a_n] \in \mathbb{R}^{n \times n}$ for attention head $h$.

\paragraph{Knowledge distillation for Transformer.} In the general framework of knowledge distillation, teacher network ($T$) with large capacity is trained in advance, and then student network ($S$) with pre-defined architecture but relatively smaller than teacher network is trained with the help of teacher's knowledge. Specifically, given the teacher parameterized by $\theta_t$, training the student parameterized by $\theta_s$ aims to minimize two objectives: i) the cross-entropy loss $\mathcal{L}^{\mathrm{CE}}$ between the output of the student network $S$ and the true label $y$ and ii) the difference of some statistics $\mathcal{L}^{\mathrm{D}}$ between teacher and student models. Overall, our goal is to minimize the following objective function:

\begin{equation}\label{loss_overall}
    \mathcal{L}(\theta_s)
    = \mathbb{E}\bigg[\mathcal{L}^{\mathrm{CE}} + \lambda \mathcal{L}^{\mathrm{D}}\Big(K^{t}(\bm{X};\theta_t), K^{s}(\bm{X};\theta_s)\Big)\bigg] \nonumber
\end{equation}
where $\lambda$ controls the relative importance between two objectives. Here, $K$ characterizes the knowledge being transferred and can vary depending on the distillation methods, and $\mathcal{L}^{\mathrm{D}}$ is a matching loss function such as $l_1$, $l_2$ or Huber loss.

Recent studies on knowledge distillation for Transformer-based BERT can also be understood in this general framework. In particular, each distillation methods of previous works are summarized in Appendix~\ref{app:prev}.

\begin{figure*}[!t]
\centering
\subfigure[]{
\includegraphics[width=.5\textwidth]{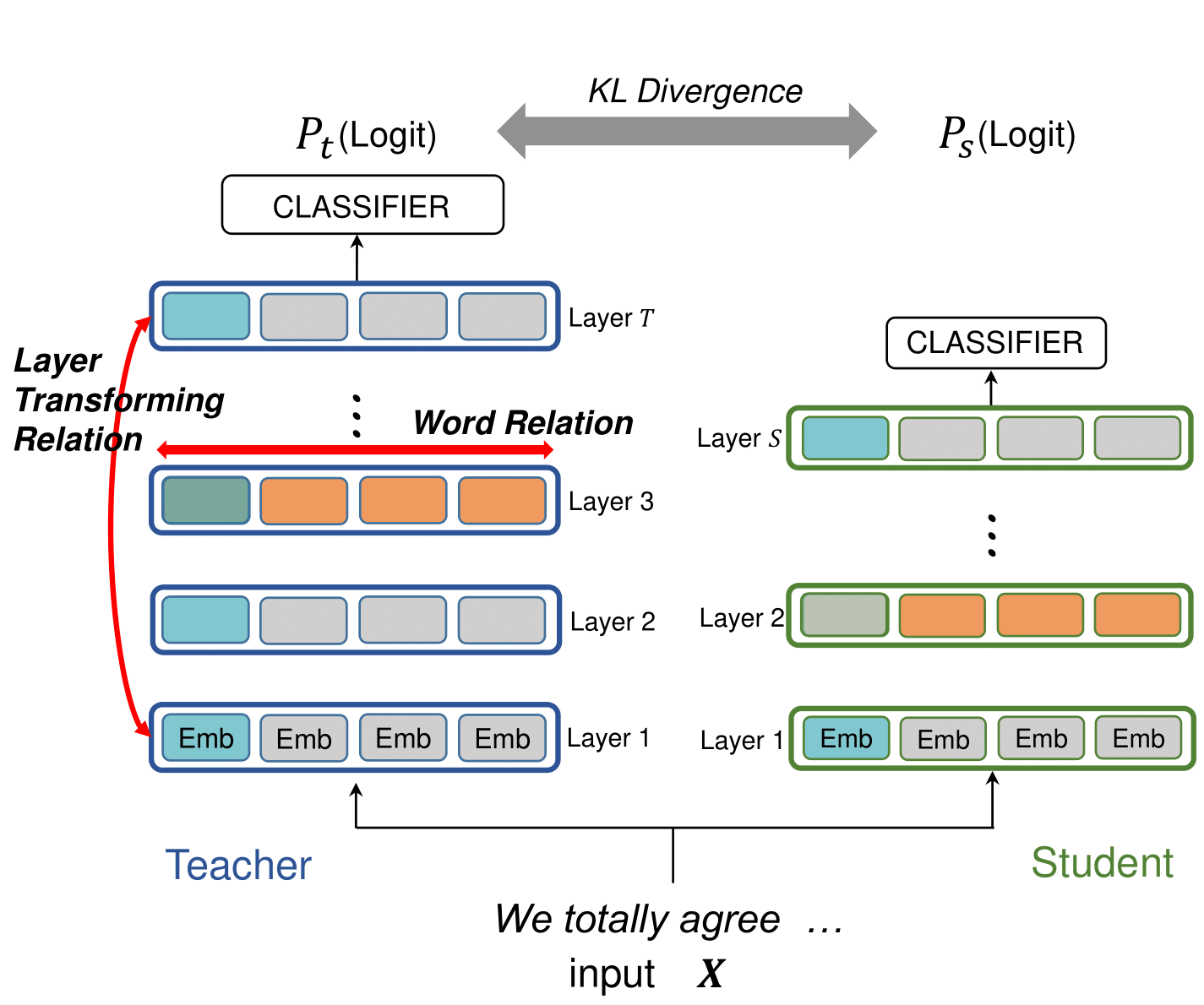}
\label{fig:main_figure1}
}
\subfigure[]{
\includegraphics[width=.46\textwidth]{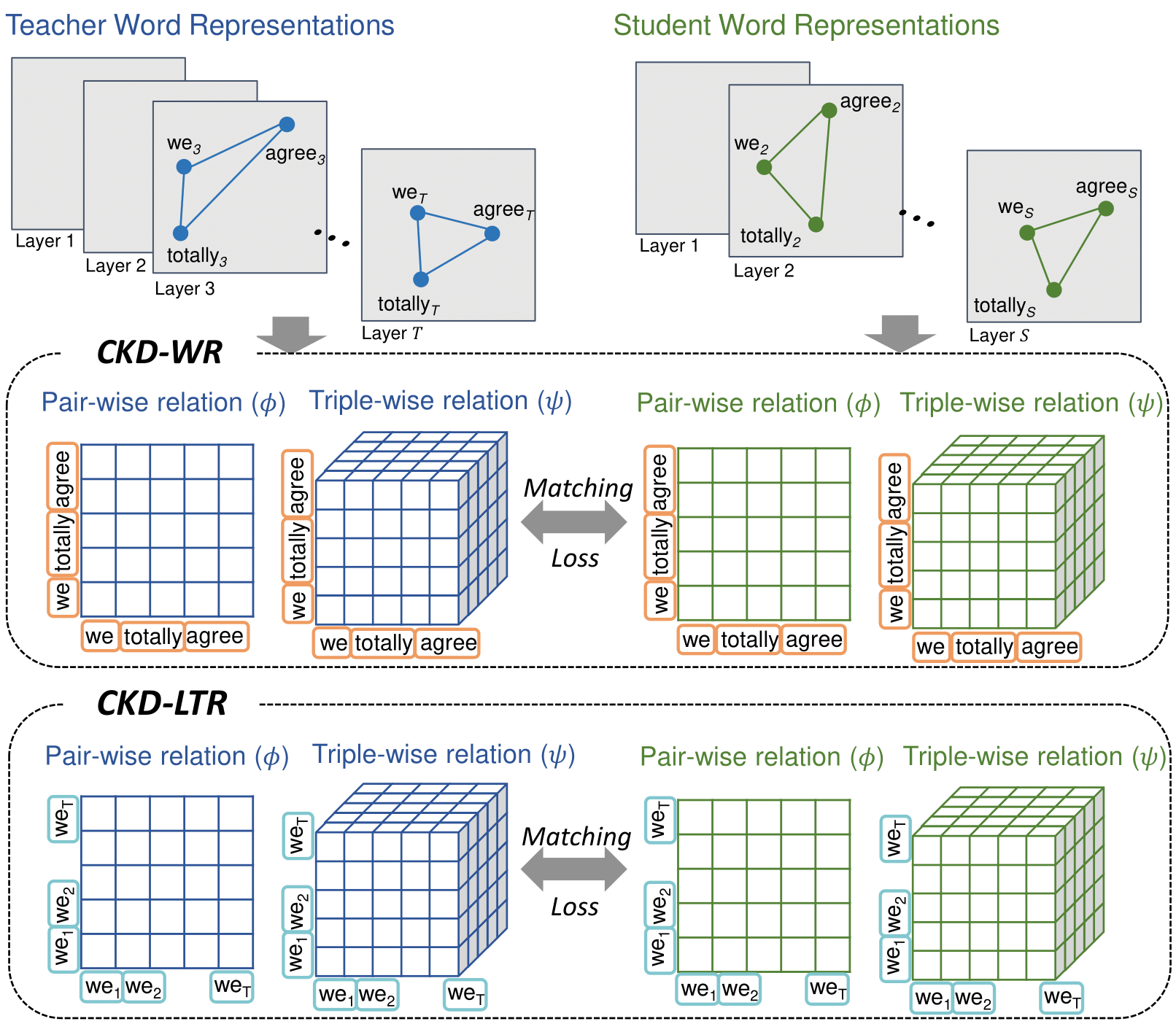}
\label{fig:main_figure2}}
\caption{\small \textbf{Overview of our contextual knowledge distillation.} 
(a) In the teacher-student framework, we define the two contextual knowledge, word relation and layer transforming relation which are the statistics of relation across the words from the same layer~(\textbf{\textcolor{ourorange}{orange})} and across the layers for the same word~(\textbf{\textcolor{ourgreen}{turquoise}}), respectively.
(b) Given the pair-wise and triple-wise relationships of WR and LTR from teacher and student, we define the objective as matching loss between them.
}
\label{fig:framework}
\end{figure*}

\section{Contextual Knowledge Distillation}\label{sec:ckd}
We now present our distillation objective that transfers the \emph{structural} or \emph{contextual} knowledge which is defined based on the distribution of word representations. Unlike previous methods distilling each word separately, our method transfers the information contained in relationships between words or between layers, and provides a more flexible way of constructing embedding space than directly matching representations. The overall structure of our method is illustrated in Figure~\ref{fig:main_figure1}. Specifically, we design two key concepts of contextual knowledge from language models: \emph{Word Relation}-based and \emph{Layer Transforming Relation}-based contextual knowledge, as shown in Figure~\ref{fig:main_figure2}.

\subsection{Word Relation~(WR)-based Contextual Knowledge Distillation}\label{ssec:wr_ckd}
Inspired by previous studies suggesting that neural networks can successfully capture contextual relationships across words \citep{manning, glove, wor2vec}, WR-based CKD aims to distill the contextual knowledge contained in the relationships across words at certain layer. The ``relationship'' across a set of words can be defined in a variety of different ways. Our work focuses on defining it as the sum of \emph{pair-wise} and \emph{triple-wise} relationships. Specifically, for each input $\bm{X}$ with $n$ words, let $\bm{R_l}=[r_{l,1},\cdots r_{l,n}]$ be the word representations at layer $l$ from the language model (it could be teacher or student), as described in Section \ref{sec:setup}. Then, the objective of WR-based CKD is to minimize the following loss: 
\begin{align}\label{EqnCKD-WR}
    &\mathcal{L}_{\mathrm{CKD-WR}} = \!\! \sum_{(i,j) \in \chi^2} \!\! w_{ij} \ \mathcal{L}^{D} \Big( \phi(r_i^s,r_j^s), \phi(r_i^t,r_j^t) \Big) \nonumber \\
    &+ \lambda_{\mathrm{WR}} \!\!\!\!\! \sum_{(i,j,k) \in \chi^3} \!\!\!\! w_{ijk} \ \mathcal{L}^{D} \Big( \psi(r_i^s,r_j^s,r_k^s), \psi(r_i^t,r_j^t,r_k^t) \Big) 
\end{align}
where $\chi=\{1,\hdots,n\}$. The function $\phi$ and $\psi$ define the pair-wise and triple-wise relationships, respectively and $\lambda_{\mathrm{WR}}$ adjust the scales of  two losses. Here, we suppress the layer index $l$ for clarity, but the distillation loss for the entire network is simply summed for all layers. Since not all terms in \myeqref{EqnCKD-WR} are equally important in defining contextual knowledge, we introduce the weight values $w_{ij}$ and $w_{ijk}$ to control the weight of how important each pair-wise and triple-wise term is. Determining the values of these weight values is open as an implementation issue, but it can be determined by the locality of words (i.e. $w_{ij} = 1$ if $|i-j| \leq \delta$ and 0, otherwise), or by attention information $\bm{A}$ to focus only on relationship between related words. In this work, we use the locality of words as weight values.

While functions $\phi$ and $\psi$ defining pair-wise and triple-wise relationship also have various possibilities, the simplest choices are to use the distance between two words for pair-wise $\phi$ and the angle by three words for triple-wise $\psi$, respectively. 

\paragraph{Pair-wise $\phi$ via distance.} Given a pair of word representations~($r_i, r_j$) from the same layer, $\phi(r_i,r_j)$ could be defined as cosine distance: $\cos{(r_i,r_j)}$ or $l_2$ distance: $\|r_i - r_j \|_2$. 

\paragraph{Triple-wise $\psi$ via angle.} Triple-wise relation captures higher-order structure and provides more flexibility in constructing contextual knowledge. One of the simplest forms for $\psi$ is the angle, which is calculated as

\begin{align}\label{eq:wr_angle}
    \psi(r_i,r_j,r_k) & = \cos \angle(r_{i}, r_{j}, r_{k}) \nonumber \\
    & = \bigg\langle \frac{r_{i} - r_{j}}{\big\|r_{i} - r_{j} \big\|_2 } , \frac{r_{k} - r_{j}} {\big\|r_{k} - r_{j} \big\|_2 }\bigg\rangle
\end{align}
where $\langle\cdot,\cdot\rangle$ denotes the dot product between two vectors. 

Despite its simple form, efficiently computing the angles in \myeqref{eq:wr_angle} for all possible triples out of $n$ words requires storing all relative representations~$(r_i - r_j)$ in a $(n,n,d_r)$ tensor\footnote{From the equation $\|r_i - r_j \|_2^2 = \|r_i\|^2_2 + \|r_j\|^2_2 - 2 \langle r_i,r_j \rangle$, computing the pair-wise distance with the right hand side of equation requires no additional memory cost.}. This incurs an additional memory cost of $\mathcal{O}(n^2 d_r)$. In this case, using locality for $w_{ijk}$ in \myeqref{EqnCKD-WR} mentioned above can be helpful; 
by considering only the triples within a distance of $\delta$ from $r_j$, the additional memory space required for efficient computation is $\mathcal{O}(\delta n d_r)$, which is beneficial for $\delta \ll n$. It also reduces the computation complexity of computing triple-wise relation from $\mathcal{O}(n^3 d_r)$ to $\mathcal{O}(\delta^2 n d_r)$. Moreover, we show that measuring angles in local window is helpful in the performance in the experimental section.

\subsection{Layer Transforming Relation~(LTR) -based Contextual Knowledge Distillation} 
The second structural knowledge that we propose to capture is on \emph{``how each word is transformed as it passes through the layers"}. Transformer-based language models are composed of a stack of identical layers and thus generate a set of representations for each word, one for each layer, with more abstract concept in the higher hierarchy. Hence, LTR-based CKD aims to distill the knowledge of how \emph{each} word develops into more abstract concept within the hierarchy. Toward this, given a set of representations for a single word in $L$ layers, $[r_{1,w}^s, \cdots, r_{L,w}^s]$ for student and $[r_{1,w}^t, \cdots, r_{L,w}^t]$ for teacher (Here we abuse the notation and $\{1,\hdots,L\}$ is not necessarily the entire layers of student or teacher. It is the index set of layers which is defined in alignment strategy; this time, we will suppress the word index below), the objective of LTR-based CKD is to minimize the following loss:

\begin{align}\label{EqnCKD-LTR}
    & \mathcal{L}_{\mathrm{CKD-LTR}} = \!\!\! \sum_{(l,m) \in \rho^2} \!\!\!\! w_{lm} \ \mathcal{L}^{D} \Big( \phi(r_l^s,r_m^s), \phi(r_l^t,r_m^t) \Big) \nonumber\\
    &+ \lambda_{\mathrm{LTR}} \!\!\!\!\!\!\!\! \sum_{(l,m,o) \in \rho^3} \!\!\!\!\!\!\! w_{lmo} \ \mathcal{L}^{D} \Big( \psi(r_l^s,r_m^s,r_o^s), \psi(r_l^t,r_m^t,r_o^t) \Big)
\end{align}
where $\rho=\{1,\hdots,L\}$ and $\lambda_{\mathrm{LTR}}$ again adjust the scales of two losses. Here, the composition of \myeqref{EqnCKD-LTR} is the same as \myeqref{EqnCKD-WR}, but only the objects for which the relationships are captured have been changed from word representations in one layer to representations for a single word in layers.
That is, the relationships among representations for a word in different layers can be defined from distance or angle as in \myeqref{eq:wr_angle}: $\phi(r_l,r_m) =  \cos(r_{l}, r_{m})$ or $\|r_l - r_m \|_2$ and $\psi(r_l,r_m,r_o) = \langle \frac{r_{l} - r_{m}}{\|r_{l} - r_{m} \|_2 } , \frac{r_{o} - r_{m}} {\|r_{o} - r_{m} \|_2 }\rangle$.

\paragraph{Alignment strategy.} When the numbers of layers of teacher and student are different, it is important to determine which layer of the student learns information from which layer of the teacher. Previous works~\citep{pkd, tinybert} resolved this \emph{alignment} issue via the \emph{uniform (i.e. skip) strategy} and demonstrated its effectiveness experimentally. 
For $L_t$-layered teacher and $L_s$-layered student, the layer matching function $f$ is defined as 
\begin{align*}
    f(\mathrm{step_s} \times t) = \mathrm{step_t} \times t, \quad \text{for } t = 0,\hdots, g
\end{align*}
where $g$ is the greatest common divisor of $L_t$ and $L_s$, $\mathrm{step_t} = L_t / g$ and $\mathrm{step_s} = L_s / g$.

\paragraph{Overall training objective.}
The distillation objective aims to supervise the student network with the help of teacher's knowledge. Multiple distillation loss functions can be used during training, either alone or together. We combine the proposed CKD with class probability matching~\citep{hinton2015distilling} as an additional term. In that case, our overall distillation objective is as follows:
\begin{equation}
    \mathcal{L} = \mathcal{L}_{\textit{logit}}^\mathrm{D} + \lambda_{\mathrm{CKD}} \Big( \mathcal{L}_{\mathrm{CKD-LTR}} + \mathcal{L}_{\mathrm{CKD-WR}} \Big) \nonumber
\end{equation}
where $\lambda_{\mathrm{CKD}}$ is a tunable parameter to balance the loss terms. 

\begin{table*}[t]
\caption{\small Comparisons for \textbf{task-agnostic} distillation. For the task-agnostic distillation comparison, we do not use task-specific distillation for a fair comparison. The results of TinyBERT and Truncated BERT are ones reported in \citet{minilm}. Other results are as reported by their authors. We exclude BERT-of-Theseus since the authors do not consider task-agnostic distillation. Results of development set are averaged over 4 runs. ``-" indicates the result is not reported in the original papers and the trained model is not released. $\dagger$ marks our runs with the officially released model by the authors.}
\begin{center}
\begin{small}
\begin{adjustbox}{max width=\textwidth}
\begin{tabular}{l|c|cccccccc|c}
\toprule
\multirow{2}{*}{Model} & \multirow{2}{*}{\#Params} & CoLA & MNLI-(m/-mm) & SST-2 & QNLI & MRPC & QQP & RTE & STS-B & \multirow{2}{*}{Avg} \\
                       & & (Mcc)& (Acc)         & (Acc) & (Acc)& (F1) &(Acc)&(Acc)&(Spear) & \\
\hline
BERT$_{\text{BASE}}$ (Teacher)& 110M & 60.4 & 84.8/84.6 &94.0 & 91.8 & 90.3 & 91.4 & 70.4& 89.5 & 84.1 \\
\hdashline

Truncated BERT~\cite{pkd} & 67.5M & 41.4 & 81.2/- & 90.8 & 87.9 & 82.7 & 90.4 & 65.5 & - & - \\ 
BERT$_\text{Small}$~\cite{pd}& 67.5M & $\text{47.1}^{\dagger}$ & 81.1/81.7 & 91.1 & 87.8 & 87.9 & 90.0 & 63.0 & $\text{87.5}^{\dagger}$ & 79.7 \\

TinyBERT~\cite{tinybert} & 67.5M & 42.8 & \textbf{83.5}/$\text{83.2}^{\dagger}$ & 91.6 & 90.5 & 88.4 & 90.6 & \textbf{72.2} & $\text{88.5}^{\dagger}$ & 81.3 \\

\hline
\textbf{CKD} & 67.5M & \textbf{52.7} & \textbf{83.5/83.4} & \textbf{92.4} & \textbf{90.7} & \textbf{89.1} & \textbf{90.8}& 70.1 & \textbf{89.1} & \textbf{82.4} \\
\bottomrule
\end{tabular}
\end{adjustbox}
\end{small}
\end{center}
\label{tb:task_agnostic}
\end{table*}

\begin{table*}[t]
\caption{\small Comparisons for \textbf{task-specific} distillation. For a fair comparison, all students are 6/768 BERT models, distilled by BERT$_{\text{BASE}}$ (12/768) teachers. Other results except for TinyBERT and PKD are as reported by their authors. Results of development set are averaged over 4 runs. ``-" indicates the result is not reported. Average score is computed excluding the MNLI-mm accuracy.}
\begin{center}
\begin{small}
\begin{adjustbox}{max width=\textwidth}
\begin{tabular}{l|c|cccccccc|c}
\toprule
\multirow{2}{*}{Model} & \multirow{2}{*}{\#Params} & CoLA & MNLI-(m/-mm) & SST-2 & QNLI & MRPC & QQP & RTE & STS-B & \multirow{2}{*}{Avg} \\
                        & & (Mcc)& (Acc)         & (Acc) & (Acc)& (F1) &(Acc)&(Acc)&(Spear) & \\
\hline
BERT$_{\text{BASE}}$ (Teacher)& 110M &60.4 & 84.8/84.6 &94.0 & 91.8 & 90.3 & 91.4 & 70.4& 89.5 & 84.1 \\
\hdashline
PD~\cite{pd} & 67.5M & - & 82.5/83.4 & 91.1 & 89.4 & 89.4 & 90.7 & 66.7& - & - \\
PKD~\cite{pkd} & 67.5M & 45.5 & 81.3/- & 91.3 & 88.4 & 85.7 & 88.4 & 66.5 & 86.2 & 79.2 \\
TinyBERT \cite{tinybert} & 67.5M &53.8 & 83.1/83.4 & 92.3 & 89.9 & 88.8 & 90.5 & 66.9 & 88.3 & 81.7 \\
BERT-of-Theseus \cite{berttheseus} & 67.5M & 51.1 & 82.3/- & 91.5 & 89.5 & 89.0 &  89.6 & \textbf{68.2} & 88.7 & 81.2 \\
\hline
\textbf{CKD} & 67.5M & \textbf{55.1} & \textbf{83.6}/\textbf{84.1} & \textbf{93.0} & \textbf{90.5} & \textbf{89.6} & \textbf{91.2} & 67.3 & \textbf{89.0} & \textbf{82.4}\\
\bottomrule
\end{tabular}
\end{adjustbox}
\end{small}
\end{center}
\label{tb:task_specific}
\end{table*}

\begin{table}[t]
\caption{\small Comparison of \textbf{task-specific} distillation on SQuAD dataset. The results of baselines and ours are reported by performing distillation with their objectives on the top of pre-trained 6-layer BERT (6/768)~\cite{pd}.}
\vspace{-0.1cm}
\begin{center}
\begin{small}
\begin{tabular}{l|c|c c}
\toprule
\multirow{2}{*}{Model} & \multirow{2}{*}{\#Params} & \multicolumn{2}{c}{SQuAD 1.1v} \\
& & EM & F1 \\
\hline
BERT$_{\text{BASE}}$ (Teacher) & 110M & 81.3 & 88.6 \\
\hline
PKD\cite{pkd} & 67.5M & 77.1 & 85.3 \\
PD\cite{pd} & 67.5M & 80.1 & 87.0 \\
TinyBERT\cite{tinybert} & 67.5M & 80.4 & 87.2 \\
\hline
\textbf{CKD} & 67.5M & \textbf{81.8} & \textbf{88.7}\\
\bottomrule
\end{tabular}
\end{small}
\end{center}
\label{tb:task_specific_squad}
\vspace{-0.2cm}
\end{table}

\subsection{Architectural Constraints in Distillation Objectives}\label{ssec:constrains}

State-of-the-art knowledge distillation objectives commonly used come with constraints in designing student networks since they directly match some parts of the teacher and student networks such as attention matrices or word representations. For example, DistilBERT~\cite{distilbert} and PKD~\cite{pkd} match each word representation independently using their cosine similarities, $\sum_{i=1}^{n} \cos(r^t_{l,i}, r^s_{l,i})$, hence the embedding size of student network should follow that of given teacher network. Similarly, TinyBERT~\cite{tinybert} and MINI-LM~\cite{minilm} match the attention matrices via $\sum_{h=1}^{H} \mathbb{KL}(\bm A_{l,h}^t, \bm A_{l,h}^s)$. Therefore, we should have the same number of attention heads for teacher and student networks (see Appendix \ref{app:prev} for more details on diverse distillation objectives). 

In addition to the advantage of distilling contextual information, our CKD method has the advantage of being able to select the student network's structure more freely without the restrictions that appear in existing KD methods. This is because CKD matches the pair-wise or triple-wise relationships of words from \emph{arbitrary} networks (student and teacher), as shown in Eq. \eqref{EqnCKD-WR}, so it is always possible to match the information of the same dimension without being directly affected by the structure. Thanks to this advantage, in the experimental section, we show that CKD can further improve the performance of recently proposed DynaBERT \citep{dynabert} that involves flexible architectural changes in the training phase. 

\section{Experiments}
We conduct task-agnostic and task-specific distillation experiments to elaborately compare our CKD with baseline distillation objectives. We then report on the performance gains achieved by our method for BERT architectures of various sizes and inserting our objective for training DynaBERT which can run at adaptive width and depth through pruning the attention heads or layers. Finally, we analyze the effect of each component in our CKD and the impact of leveraging locality $\delta$ for $w_{ijk}$ in~\myeqref{EqnCKD-WR}. 
\paragraph{Dataset.} For task-agnostic distillation which compresses a large pre-trained language model into a small language model on the pre-training stage, we use a document of English Wikipedia. For evaluating the compressed language model on the pre-training stage and task-specific distillation, we use the GLUE benchmark~\cite{glue} which consists of nine diverse sentence-level classification tasks and SQuAD~\cite{squad}.

\paragraph{Setup.} For task-agnostic distillation, we use the original BERT without fine-tuning as the teacher. Then, we perform the distillation on the student where the model size is pre-defined. We perform distillation using our proposed CKD objective with class probability matching of masked language modeling for 3 epochs while task-agnostic distillation following the \citet{tinybert} and keep other hyperparameters the same as BERT pre-training~\cite{bert}. For task-specific distillation, we experiment with our CKD on top of pre-trained BERT models of various sizes which are released for research in institutions with fewer computational resources\footnote{https://github.com/google-research/bert}~\cite{pd}. For the importance weight of each pair-wise and triple-wise terms, we leverage the locality of words, in that $w_{ij} = 1$ if $|i-j| \leq \delta$ and 0, otherwise. For this, we select the $\delta$ in (10-21). More details including hyperparameters are provided in Appendix \ref{app:exp_setting}. The code to reproduce the experimental results is available at
\href{https://github.com/GeondoPark/CKD}{https://github.com/GeondoPark/CKD}.

\begin{figure*}[t]
\centering
\includegraphics[width=1\linewidth]{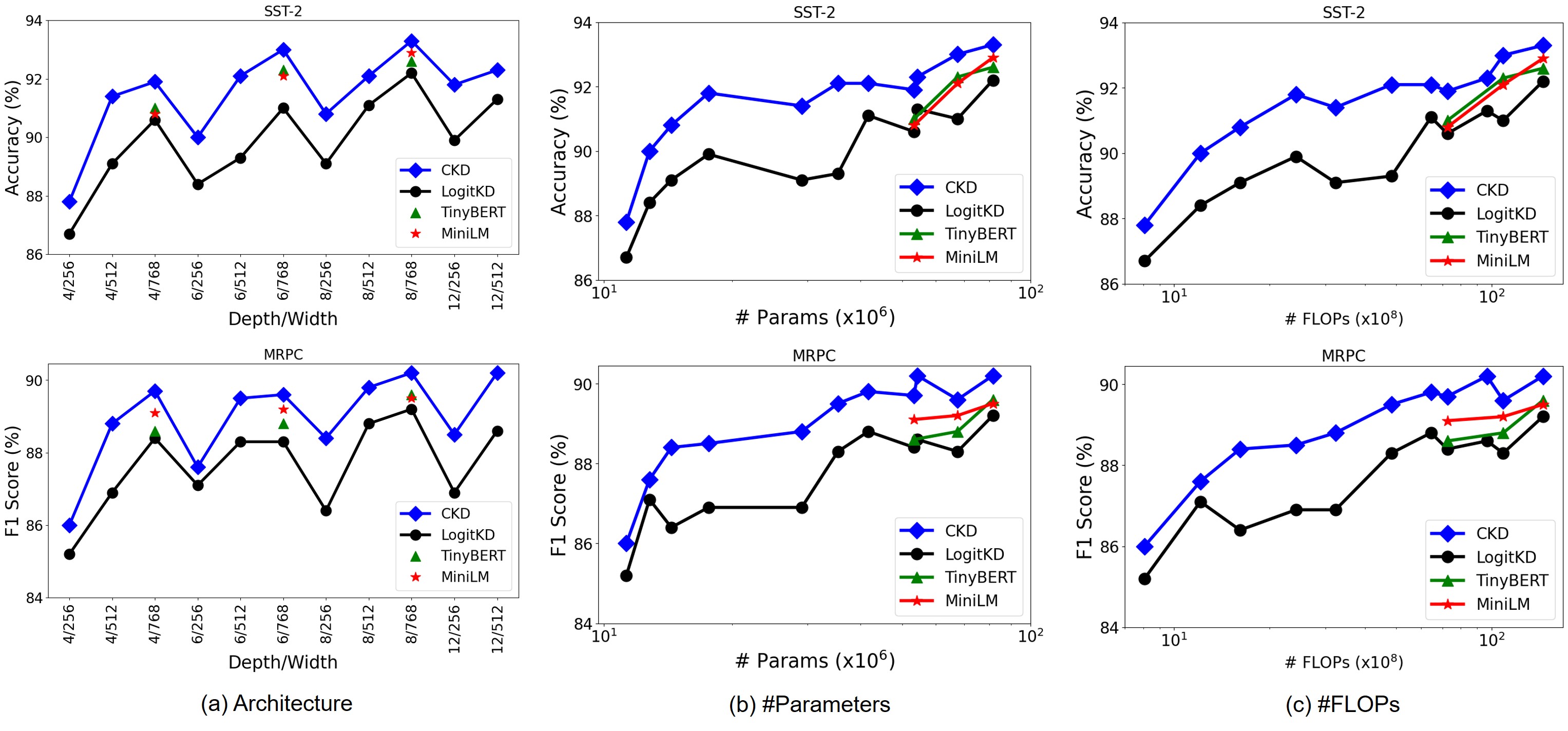}
\caption{\small Task specific distillation on various sizes of models. We consider diverse cases by changing (a) the network structures, (b) the number of parameters and (c) the number of FLOPs. All results are averaged over 4 runs on the development set.} 
\label{fig:variation_size}
\end{figure*}

\begin{figure*}[t]
\centering
\includegraphics[width=1\linewidth]{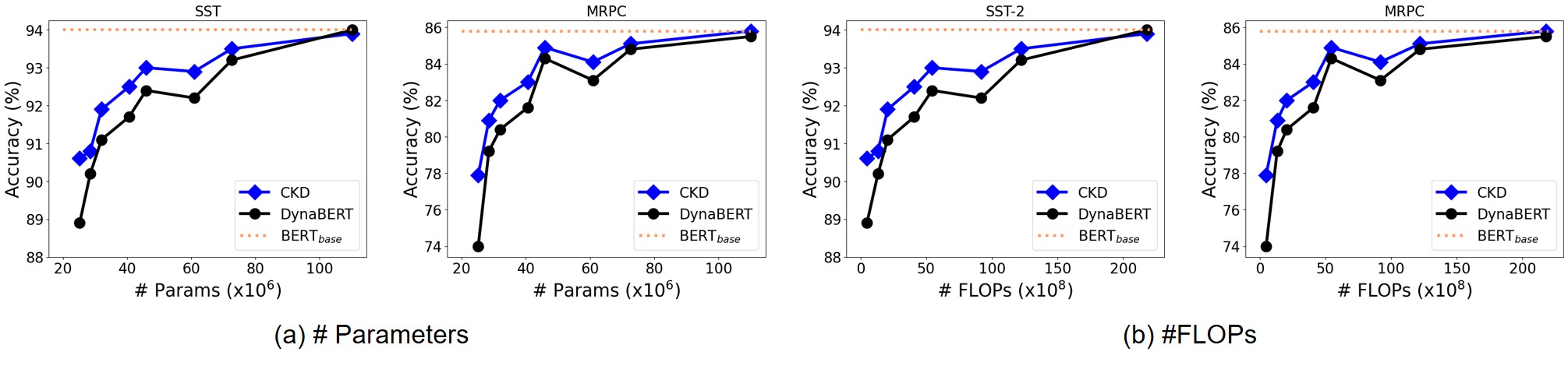}
\caption{\small Boosting the performance of DynaBERT via training with CKD. Comparison between the original DynaBERT and CKD-augmented DynaBERT according to (a) the number of parameters and (b) the number of FLOPs. The results are averaged over 4 runs on the development set.}
\label{fig:dynabert}
\end{figure*}

\subsection{Main Results} \label{ssec:main_results}
To verify the effectiveness of our CKD objective, we compare the performance with previous distillation methods for BERT compression including task-agnostic and task-specific distillation. Following the standard setup in baselines, we use the BERT$_{\text{BASE}}$~(12/768)\footnote{In notation~$(a/b)$, $a$ means the number of layers and $b$ denotes a hidden size in intermediate layers. The number of attention heads is defined as $b/64$.} as the teacher and 6-layer BERT~(6/768) as the student network. Therefore, the student models used in all baselines and ours have the same number of parameters (67.5M) and inference FLOPs (10878M) and time.

\paragraph{Task-agnostic Distillation.} We compare with three baselines: 1) Truncated BERT which drop top 6 layers from BERT$_{\text{base}}$ proposed in PKD~\cite{pkd}, 2) BERT$_{\text{small}}$ which trained using the Masked LM objectives provided in PD~\cite{pd}, 3) TinyBERT~\cite{tinybert} which propose the individual word representation and attention map matching. Since MobileBERT~\cite{mobilebert} use the specifically designed teacher and student network which have 24-layers with an inverted bottleneck structure, we do not compare with. DistilBERT~\cite{distilbert} and MINI-LM~\cite{minilm} use the additional BookCorpus dataset which is no longer publicly available. Moreover, the authors do not release the code, making it hard to reproduce. Thus we do not compare in the main paper for a fair comparison. The comparisons with those methods are available in Appendix~\ref{app:more_com}. Results of task-agnostic distillation on GLUE dev sets are presented Table~\ref{tb:task_agnostic}. The result shows that CKD surpasses all baselines. Comparing with TinyBERT which transfers the knowledge of individual representations, CKD outdoes in all scores except for the RTE. These results empirically demonstrate that distribution-based knowledge works better than individual representation knowledge. 

\begin{figure*}[t]
\begin{minipage}[h]{0.64\linewidth}
\captionsetup{labelfont=footnotesize}
\centering
\captionof{table}{\small Ablation study about the impact of each component of CKD. '- *' denotes to eliminate *, the component of CKD.}
{\small
\begin{adjustbox}{max width=0.98\textwidth}
\begin{tabular}{l|cccccc}
\toprule
\multirow{2}{*}{Objectives} & MNLI-(m/-mm) & SST-2 & QNLI & MRPC & QQP & STS-B \\
& (Acc) & (Acc) & (Acc) & (F1) & (Acc) & (Spear) \\
\hline
CKD & \textbf{80.7}/\textbf{80.8} & \textbf{91.4} & \textbf{88.1} & \textbf{88.8} & \textbf{90.3} & \textbf{87.9} \\
\hline 
 $\quad$- WR & 80.1/80.6 & 90.6 & 87.5 & 88.5 & 89.7 & 87.5 \\
 $\quad$- LTR & 79.9/80.3 & 91.1 & 87.8 & 88.3 & \textbf{90.3} & 87.6\\ 
 $\quad$- WR - LTR & 79.2/79.9 & 89.1 & 87.4 & 88.1 & 89.2 & 86.8 \\ 
\bottomrule
\end{tabular}
\end{adjustbox}
}
\label{tb:ablation1}
\end{minipage} \hfill
\begin{minipage}[h]{0.35\linewidth}
{\small
\begin{adjustbox}{max width=0.98\textwidth}
\includegraphics[width=1\linewidth]{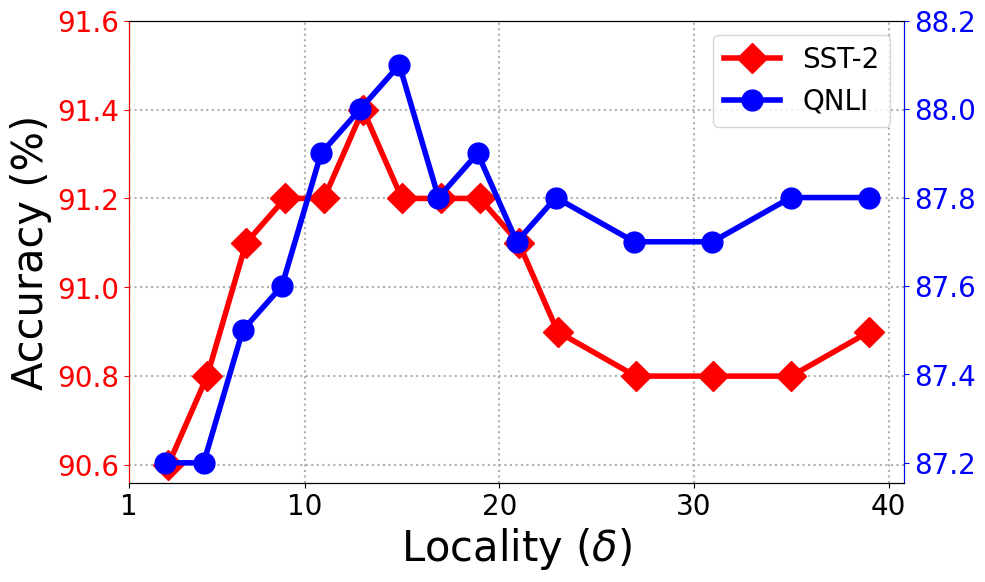}
\label{fig:locality}
\end{adjustbox}
}
\vspace{-0.8cm}
\captionof{figure}{\small Effect of local window size.}
\label{fig:ablation2}
\end{minipage}
\end{figure*}

\paragraph{Task-specific Distillation.} Here, we compare with four baselines that do not perform distillation in the pre-training: 1) PD~\cite{pd} which do pre-training with Masked LM and distills with Logit KD in task-specific fine-tuning process. 2) PKD~\cite{pkd} which uses only 6 layers below BERT$_\text{base}$, and distillation is also performed only in task-specific fine-tuning. The GLUE results on dev sets of PKD are taken from \citep{berttheseus}. 3) TinyBERT~\cite{tinybert}. For the TinyBERT, we also perform distillation only in the task-specific fine-tuning with their objectives on the top of the pre-trained model provided by \citet{pd} for a fair comparison. 4) BERT-of-Theseus~\cite{berttheseus} which learn a compact student network by replacing the teacher layers in a fine-tuning stage. Results of task-specific distillation on GLUE dev sets and SQuAD datasets are presented in Table~\ref{tb:task_specific} and~\ref{tb:task_specific_squad}, respectively. Note that briefly, the CKD also outperforms all baselines for all GLUE datasets and SQuAD dataset except for RTE for task-specific distillation, convincingly verifying its effectiveness. These results consistently support that contextual knowledge works better than other distillation knowledge.

\subsection{Effect of CKD on various sizes of models}
For the knowledge distillation with the purpose of network compression, it is essential to work well in more resource-scarce environments. To this end, we further evaluate our method on various sizes of architectures. For this experiment, we perform distillation on a task-specific training process on top of various size pre-trained models provided by \citet{pd}. We compare CKD with three baselines: 1) LogitKD objective used by~\citet{distilbert, pd}. 2) TinyBERT~\cite{tinybert} objective which includes individual word representations and attention matrix matching. 3) MINI-LM~\cite{minilm} objective which includes attention matrix and value-value relation matching. We implement the baselines and runs for task-specific distillation. We note that MINI-LM and TinyBERT objective are applicable only to models (*/768) which have the same number of attention heads with the teacher model (12/768). Figure \ref{fig:variation_size} illustrate that our CKD consistently exhibits significant improvements in the performance compared LogitKD. In addition, for task-specific distillation, we show that CKD works better than all baselines on (*/768) student models. The results on more datasets are provided in Appendix~\ref{app:more_results}.

\subsection{Incorporating with DynaBERT}
DynaBERT~\cite{dynabert} is a recently proposed adaptive-size pruning method that can run at adaptive width and depth by removing the attention heads or layers. In the training phase, DynaBERT uses distillation objectives which consist of LogitKD and individual word representations matching to improve the performance. Since the CKD objective has no constraints about architecture such as embedding size or the number of attention heads, we validate the objective by replacing it with CKD. The algorithm of DynaBERT and how to insert CKD are provided in Appendix~\ref{app:detail_dyna}. To observe just how much distillation alone improves performance, we do not use data augmentation and an additional fine-tuning process. We note that objectives proposed in MINI-LM~\cite{minilm} and TinyBERT~\cite{tinybert} cannot be directly applied due to constraints of the number of attention heads. As illustrated in Figure~\ref{fig:dynabert}, CKD consistently outperforms the original DynaBERT on dynamic model sizes, supporting the claim that distribution-based knowledge is more helpful than individual word representation knowledge. The results on more datasets are provided in Appendix~\ref{app:more_results}. 

\subsection{Ablation Studies}
We provide additional ablation studies to analyze the impact of each component of the CKD and the introduced locality~($w_{i,j}=\delta$) in~\myeqref{EqnCKD-WR} as the weight of how important each pair-wise and triple-wise term is. For these studies, we fix the student network with 4-layer BERT~(4/512) and report the results as an average of over 4 runs on the development set. 

\paragraph{Impact of each component of CKD.}
The proposed CKD transfers the word relation based and layer transforming relation based contextual knowledge. To isolate the impact on them, we experiment successively removing each piece of our objective. Table~\ref{tb:ablation1} summarizes the results, and we observe that WR and LTR can bring a considerable performance gain when they are applied together, verifying their individual effectiveness. 

\paragraph{Locality as the importance of relation terms.} We introduced the additional weights ($w_{ij}$, $w_{ijk}$) in \myeqref{EqnCKD-WR} for CKD-WR (and similar ones for CKD-LTR) to control the importance of each pair-wise and triple-wise term and suggested using the locality for them as one possible way. Here, we verify the effect of locality by increasing the local window size~($\delta$) on the SST-2 and QNLI datasets. The result is illustrated in Figure~\ref{fig:ablation2}. We observe that as the local window size increases, the performance improves, but after some point, the performance is degenerated. From this ablation study, we set the window size ($\delta$) between 10-21. 
\vspace{-0.1cm}
\section{Conclusion}

We proposed a novel distillation strategy that leverages contextual information efficiently based on word relation and layer transforming relation. To our knowledge, we are the first to apply this contextual knowledge which is studied to interpret the language models. Through various experiments, we show not only that CKD outperforms the state-of-the-art distillation methods but also the possibility that our method boosts the performance of other compression methods.


\section*{Acknowledgement}
This work was supported by the National Research Foundation of Korea (NRF) grants (2018R1A5A1059921, 2019R1C1C1009192) and Institute of Information \& Communications Technology Planning \& Evaluation (IITP) grants (No.2017-0-01779, A machine learning and statistical inference framework for explainable artificial intelligence, No.2019-0-01371, Development of brain-inspired AI with human-like intelligence, No.2019-0-00075, Artificial Intelligence Graduate School Program (KAIST)) funded by the Korea government (MSIT). 

\bibliography{anthology,custom}
\bibliographystyle{acl_natbib}
\clearpage
\appendix
\begin{table*}[t]
\caption{Overview of distillation objectives used for language model compression and their constraint on architecture. $\mathbb{S}_k$ means scaled softmax function across the $k$th-dimension.}
\vspace{-0.5cm}
\begin{center}
\begin{footnotesize}
\begin{adjustbox}{max width=0.9\textwidth}
\begin{tabular}{l|c|c}
\toprule
 & \textbf{Knowledge Distillation Objectives} & \textbf{Constraint} \\
\hline
DistilBERT~\cite{distilbert}& $\mathlarger{\sum}_{i=1}^{n} \cos(r^t_{l,i}, r^s_{l,i})$, $\mathcal{L}^{\mathrm{D}}_\textit{Logit}$ & Embedding size \\
\hline
PKD~\cite{pkd} &  $ \mathlarger{\sum}_{i=1}^{n} \Big[\texttt{MSE}(\mfrac{r^{t}_{l,i}}{\|r^{t}_{l,i} \|_{2}} - \mfrac{r^{s}_{l,i}}{ \|r^{s}_{l,i} \|_{2}}) \Big]$, $\mathcal{L}^{\mathrm{D}}_\textit{Logit}$ & Embedding size \\
\hline
TinyBERT~\cite{tinybert} & $\mathlarger{\sum}_{i=1}^{n} \Big[\texttt{MSE}(r^{t}_{l,i} - W_{r} r^{s}_{l,i})\Big]$, $  \mathlarger{\sum}_{h=1}^{H}\Big[\texttt{MSE}{(\bm A^{t}_{l,h}} - \bm A^{s}_{l,h})\Big]$, $\mathcal{L}^{\mathrm{D}}_\textit{Logit}$ & Attention head \\ 
\hline
Mobile-BERT~\cite{mobilebert} & $\mathlarger{\sum}_{i=1}^{n} \Big[\texttt{MSE}(r^{t}_{l,i} - r^{s}_{l,i})\Big]$, $\mathlarger{\sum}_{h=1}^{H} \Big[ \mathbb{KL}\big(\bm A_{l,h}^t, \bm A_{l,h}^s \big) \Big]$, $\mathcal{L}^{\mathrm{D}}_\textit{Logit}$ & \makecell{Embedding size \\ Attention head} \\
\hline
MiniLM~\cite{minilm} & $\mathlarger{\sum}_{h=1}^{H} \Big[ \mathbb{KL}\big(\bm A_{l,h}^t, \bm A_{l,h}^s \big) \Big]$ , $\mathlarger{\sum}_{h=1}^{H} \Big[ \mathbb{KL} \big( \mathbb{S}_{2} (\bm{V}_{l,h}^t \cdot {\bm{V}_{l,h}^t}^T), \mathbb{S}_{2} (\bm{V}_{l,h}^s \cdot {\bm{V}_{l,h}^s}^T \big) \Big] $ &  Attention head \\
\bottomrule
\end{tabular}
\end{adjustbox}
\label{table: previousework}
\end{footnotesize}
\end{center}
\vspace{-0.2cm}
\end{table*}

\section{Explanation of previous methods and their constraints}\label{app:prev}
Table \ref{table: previousework} present the details of knowledge distillation objectives of previous methods and their constraints.

\textbf{DistilBERT}~\cite{distilbert} uses logit distillation loss~(Logit KD), masked language modeling loss, and cosine loss between the teacher and student word representations in the learning process. The cosine loss serves to align the directions of the hidden state vectors of the teacher and student. Since the cosine of the two hidden state vectors is calculated in this process, they have the constraint that the embedding size of the teacher and the student model must be the same.

\textbf{PKD}~\cite{pkd} transfers teacher knowledge to the student with Logit KD and patient loss. The patient loss is the mean-square loss between the normalized hidden states of the teacher and student. To calculate the mean square error between the hidden states, they have a constraint that the dimensions of hidden states must be the same between teacher and student.


\textbf{TinyBERT}~\cite{tinybert} uses additional loss that matches word representations and attention matrices between the teacher and student. Although they acquire flexibility on the embedding size, using an additional parameter, since the attention matrices of the teacher and student are matched through mean square error loss, the number of attention heads of the teacher and student must be the same. 

\textbf{MobileBERT}~\cite{mobilebert} utilizes a similar objective with TinyBERT~\citep{tinybert} for task-agnostic distillation. However, since they match the hidden states with $l2$ distance and attention matrices with $\mathcal{KL}$ divergence between teacher and student, they have restrictions on the size of hidden states and the number of attention heads. 

\textbf{MiniLM}~\cite{minilm} proposes distilling the self-attention module of the last Transformer layer of the teacher. In self-attention module, they transfer attention matrices such as TinyBERT and MobileBERT and Value-Value relation matrices. Since they match the attention matrices of the teacher and student in a one-to-one correspondence, the number of attention heads of the teacher and student must be the same.

The methods introduced in Table \ref{table: previousework} have constraints by their respective knowledge distillation objectives. However, our CKD method which utilizes the relation statistics between the word representations~(hidden states) has the advantage of not having any constraints on student architecture.

\section{Details of experiment setting}\label{app:exp_setting}
This section introduces the experimental setting in detail. We implemented with PyTorch framework and huggingface's transformers package~\citep{HuggingFaces}. 

\paragraph{Task-agnostic distillation} We use the pre-trained original BERT$_\text{base}$ with masked language modeling objective as the teacher and a document of English Wikipedia as training data. We set the max sequence length to 128 and follow the preprocess and WordPiece tokenization of \citet{bert}. Then, we perform the distillation for 3 epochs. For the pre-training stage, we use the CKD objective with class probability matching of masked language modeling and keep other hyper-parameters the same as BERT pre-training~\cite{bert}.

\paragraph{Task-specific distillation} Our contextual knowledge distillation proceeds in the following order. First, from pre-trained BERT$_\text{base}$, task-specific fine-tuning is conducted to serve as a teacher. Then, prepare the pre-trained small-size architecture which serves as a student. In this case, pre-trained models of various model sizes provided by \citet{pd} are employed. Finally, task-specific distillation with our CKD is performed. 

To reduce the hyperparameters search cost, $\lambda_{\mathrm{WR}}$ in \myeqref{EqnCKD-WR} and $\lambda_{\mathrm{LTR}}$ in \myeqref{EqnCKD-LTR} are used with same value. For the importance weights introduced for pair-wise and triple-wise terms, the locality is applied only to the importance weight $w$ of the word relation (WR)-based CKD loss. The importance weight $w$ of the layer transforming relation~(LTR)-based CKD loss is set to 1. In this paper, we report the best result among the following values to find the optimal hyperparameters of each dataset:
\begin{itemize}
    \item Alpha ($\alpha$) : 0.7, 0.9
    \item Temperature ($T$) : 3, 4
    \item $\lambda_{\mathrm{WR}}$, $\lambda_{\mathrm{LTR}}$ : 1, 10, 100, 1000
    \item $\lambda_{\mathrm{CKD}}$ : 1, 10, 100, 1000
\end{itemize}
Other training configurations such as batch size, learning rate and warm up proportion are used following the BERT~\cite{bert}.

\begin{table*}[h]
\caption{\small Full comparison of task-agnostic distillation comparing our CKD against the baseline methods. For the task-agnostic distillation comparison, we do not use task-specific distillation for a fair comparison. The results of TinyBERT cited as reported by \citet{minilm}. Other results are as reported by their authors. Results of the development set are averaged over 4 runs. ``-" means the result is not reported and the trained model is not released. $\dagger$ marks our runs with the officially released model.}
\vspace{-0.3cm}
\begin{center}
\begin{small}
\begin{adjustbox}{max width=\textwidth}
\begin{tabular}{l|c|cccccccc}
\toprule
\multirow{2}{*}{Model} & \multirow{2}{*}{\#Params} & CoLA & MNLI-(m/-mm) & SST-2 & QNLI & MRPC & QQP & RTE & STS-B \\
                       & & (Mcc)& (Acc)         & (Acc) & (Acc)& (F1) &(Acc)&(Acc)&(Spear) \\
\hline
BERT$_{\text{BASE}}$ (Teacher)& 110M & 60.4 & 84.8/84.6 &94.0 & 91.8 & 90.3 & 91.4 & 70.4& 89.5 \\
\hdashline
Truncated BERT~\cite{pkd} & 67.5M & 41.4 & 81.2/- & 90.8 & 87.9 & 82.7 & 90.4 & 65.5 & - \\ 
BERT$_\text{Small}$~\cite{pd}& 67.5M & $\text{47.1}^{\dagger}$ & 81.1/81.7 & 91.1 & 87.8 & 87.9 & 90.0 & 63.0 & $\text{87.5}^{\dagger}$  \\
DistilBERT~\cite{distilbert} & 67.5M & 51.3 & 82.2/- & 91.3 & 89.2 & 87.5 & 88.5 & 59.9& 86.9 \\
TinyBERT~\cite{tinybert} & 67.5M & 42.8 & 83.5/$\text{83.2}^{\dagger}$ & 91.6 & 90.5 & 88.4 & 90.6 & \textbf{72.2} & $\text{88.5}^{\dagger}$ \\
MINI-LM~\cite{minilm} & 67.5M & 49.2 &  \textbf{84.0}/- & 92.0 & \textbf{91.0} & 88.4 & \textbf{91.0} & 71.5 & - \\
\hline
\textbf{CKD} & 67.5M & \textbf{52.7} & 83.5/\textbf{83.4} & \textbf{92.4} & 90.7 & \textbf{89.1} & 90.8 & 70.1 & \textbf{89.1}  \\
\bottomrule
\end{tabular}
\end{adjustbox}
\vspace{-0.2cm}
\end{small}
\end{center}
\label{tb:task_agnostic_full}
\end{table*}

\section{Additional comparison on task-agnostic distillation}\label{app:more_com}

We report the fair comparison of our method and baselines about the task-agnostic distillation in Section~\ref{ssec:main_results} of the main paper. Several works~\cite{distilbert, minilm} use the additional BookCorpus dataset which is no longer publicly available. Here, we present the full comparison of CKD and baselines including DistilBERT~\cite{distilbert} and MINI-LM~\cite{minilm}. As shown in Table~\ref{tb:task_agnostic_full}, even though we do not use the BookCorpus dataset, we outperform all baselines on four datasets and obtain comparable performance on the rest of the datasets.


\section{Applying CKD to DynaBERT}\label{app:detail_dyna}

In this section, we describe how we apply our CKD objective to DynaBERT~\citep{dynabert}. Training DynaBERT consists of three stages: 1) Rewire the model according to the importance and then 2) Go through the two-stage of adaptive pruning with distillation objective. Since we suppress some details of DynaBERT for clarity, refer to the paper~\cite{dynabert} for more information.

We summarize the training procedure of DynaBERT with CKD in algorithm \ref{alg:dynabert_w/CKD}. To fully exploit the capacity, more important attention heads and neurons must be shared more across the various sub-networks. Therefore, we follow phase 1 in DynaBERT to rewire the network by calculating the loss and estimating the importance of each attention head in the Multi-Head Attention~(MHA) and neuron in the Feed-Forward Network~(FFN) based on gradients. Then, they train the DynaBERT by accumulating the gradient varying the width and depth of BERT. In these stages, they utilize distillation objective which matches hidden states and logits to improve the performance. We apply our CKD at these stages by replacing their objective with CKD as shown in algorithm \ref{alg:dynabert_w/CKD}~(Blue). Since CKD has no restrictions on student’s architecture, it can be easily applied.

\begin{algorithm*}[ht]
\small
    \DontPrintSemicolon
    \SetAlgoVlined
    \SetKw{Initialize}{initialize}
    \SetKwInOut{Input}{input}
    \SetKwBlock{PhaseOne}{Phase 1: Rewire the network.}{}
    \SetKwBlock{PhaseTwo}{Phase 2: Train DynaBERT$_\text{W}$ with adaptive width.}{}
    \SetKwBlock{PhaseThree}{Phase 3: Train DynaBERT with adaptive width and depth.}{}
    \PhaseOne{
        \Input{Development set, trained BERT on downstream task.}
        Calculate the importance of attention heads and neurons with gradients.\;
        Rewire the network according to the importance.\;
    }
    \PhaseTwo{
        \Input{Training set, width multiplier list $widthList$.}
        \Initialize{a fixed teacher model and a trainable student model with rewired net.}\;
        \For{$iter=1,\dots,T_{train}$} {
            Get the logits $\bm{y}$, hidden states $\bm{R}$ from teacher model.\;
            \For{width multiplier $m_w$ in $widthList$} {
                Get the logits $\bm{y}^{(m_w)}$, hidden states $\bm{R}^{(m_w)}$ from student model.\;
                Compute distillation loss. \; 
                $\mathcal{L}_\mathrm{{DynaBERT}}=\mathrm{SCE}(\bm{y}^{(m_w)},\bm{y})+ \lambda_{1} \cdot \sum_{l=0}^{L}\mathrm{MSE}(\bm{R}_l^{(m_w)},\bm{R}_l)$\;
                \textcolor{blue}{$\mathcal{L}_{\mathrm{CKD}}=\mathrm{SCE}(\bm{y}^{(m_w)},\bm{y})+ \lambda_{1} \cdot \mathcal{L}_{\mathrm{CKD-WR}}(\bm{R}^{(m_w)}, \bm{R}) + \lambda_{2} \cdot \mathcal{L}_{\mathrm{CKD-LTR}}(\bm{R}^{(m_w)}, \bm{R}) $}\;
                Accumulate gradients $\mathcal{L}.backward()$.\; 
            }
            Update with the accumulated gradients.\;
        }
    }
    \PhaseThree{
        \Input{Training set, width multiplier list $widthList$, depth multiplier list $depthList$.}
        \Initialize{a fixed DynaBERT$_\text{W}$ as teacher model and a trainable student model with the DynaBERT$_\text{W}$.}\;
        \For{$iter=1,\dots,T_{train}$} {
            \For{width multiplier $m_w$ in $widthList$} {
                Get the logits $\bm{y}^{(m_w)}$, $\bm{R}^{(m_w)}$ from teacher model.\;
                \For{depth multiplier $m_d$ in $depthList$} {
                    Get the logits $\bm{y}^{(m_w,m_d)}$, hidden states $\bm{R}^{(m_w,m_d)}$ from student model.\;
                    Compute distillation loss. \;  $\mathcal{L}_\mathrm{{DynaBERT}}=\mathrm{SCE}(\bm{y}^{(m_w,m_d)},\bm{y}^{(m_w)})+\lambda_{1} \cdot \sum_{l,l'\in{L_S,L_T}}\mathrm{MSE}(\bm{R}_l^{(m_w, m_d)},\bm{R}_{l'}^{(m_w)})$\;
                    \textcolor{blue}{$\mathcal{L}_{\mathrm{CKD}}=\mathrm{SCE}(\bm{y}^{(m_w,m_d)},\bm{y}^{(m_w)})+\lambda_{1} \cdot \mathcal{L}_{\mathrm{CKD-WR}}(\bm{R}^{(m_w, m_d)},\bm{R}^{(m_w)})$ \;
                    $+ \lambda_{2} \cdot \mathcal{L}_{\mathrm{CKD-LTR}}(\bm{R}^{(m_w, m_d)},\bm{R}^{(m_w)}) $}\;
                    Accumulate gradients $\mathcal{L}.backward()$.\; 
                }
            }
            Update with the accumulated gradients.\;
        }
    }
    \caption{Train DynaBERT with CKD}
    \label{alg:dynabert_w/CKD}
\end{algorithm*}

\section{More Results}\label{app:more_results}

Due to space limitations in the main paper, we only report the results on a subset of GLUE datasets for experiments about the effect of model size for CKD and boosting the DynaBERT with CKD. Here, we report all datasets of GLUE except for CoLA for two experiments. We exclude the CoLA dataset since the distillation losses are not converged properly in the very small-size models.

Here, we present the results of three experiments on additional datasets in order. 1) Effect of CKD on various sizes of models. 2) Boosting the performance of DynaBERT when CKD is applied.

\begin{figure*}[t]
\centering
\includegraphics[width=1\linewidth]{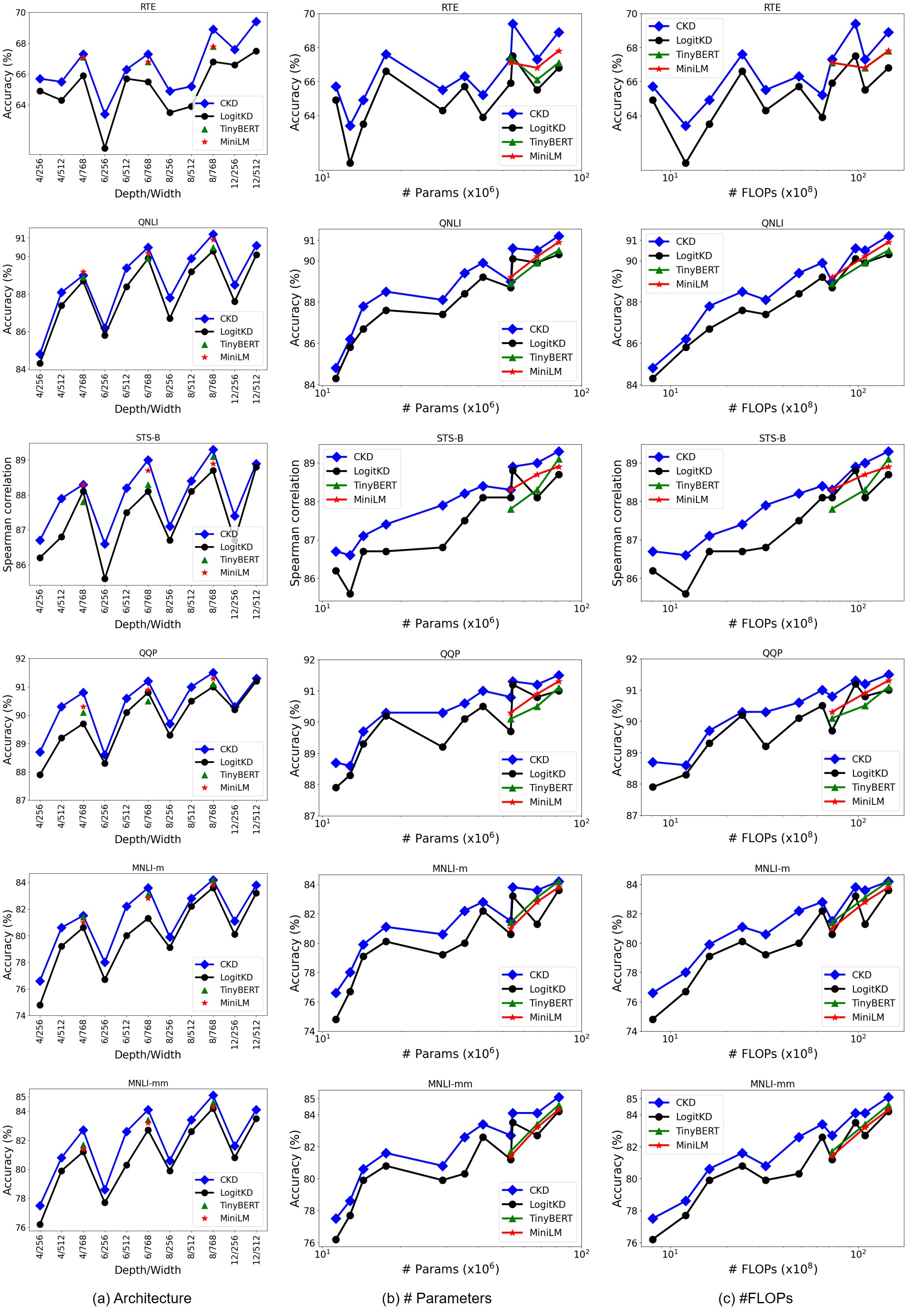}
\caption{\small The efficiency of various sizes of models for CKD compared to baselines. The performance graph according to (a) network structure (b) the number of parameters (c) the number of FLOPs. The results are averaged over 4 runs on the development set.} 
\vspace{-0.2cm}
\label{fig:app_variation_size}
\end{figure*}

\paragraph{Effect of CKD on various sizes of models.} 
Figure~\ref{fig:app_variation_size} illustrates the performance of task-specific distillation on various sizes of models. Again, we note that MINI-LM and TinyBERT objectives are applicable only to models (*/768), which have the same number of attention heads as the teacher model (12/768). As shown in Figure~\ref{fig:app_variation_size}, our CKD consistently exhibits significant improvements in the performance compared LogitKD for all model sizes. Compared to TinyBERT and MINI-LM, CKD shows higher performance on all datasets for almost all model sizes (*/768). 

\begin{figure*}[t]
\centering
\includegraphics[width=1\linewidth]{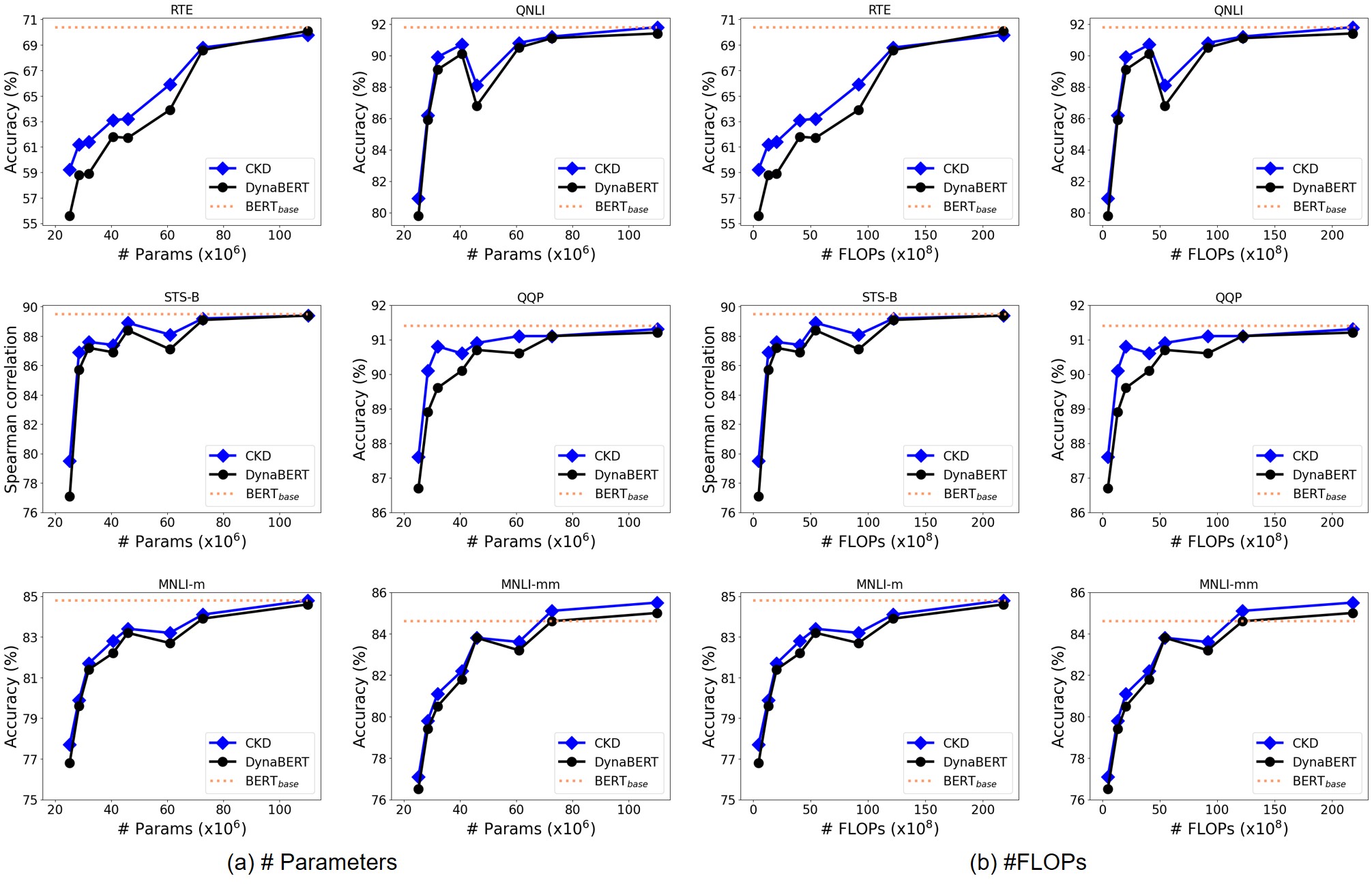}
\vspace{-0.7cm}
\caption{\small Boosting the performance of DynaBERT when CKD is applied. The performance graph for comparison of original DynaBERT and CKD according to (a) the number of parameters and (b) the number of FLOPs. The results are averaged over 4 runs on the development set.}
\label{fig:app_dynabert}
\end{figure*}

\paragraph{Incorporating with DynaBERT} 
Figure~\ref{fig:app_dynabert} shows the performance of the original DynaBERT and when CKD is applied. As illustrated in Figure~\ref{fig:app_dynabert}, CKD further improves the original DynaBERT on dynamic width and depth size, convincingly verifying its effectiveness. The results also present the possibility that our method boosts the performance of other compression methods. 

\end{document}